\documentclass[10pt,twocolumn,letterpaper]{article}

\usepackage{cvpr}              

\usepackage{multicol}
\usepackage{multirow}
\usepackage{indentfirst}
\usepackage{tabularx}
\usepackage{caption}
\usepackage[dvipsnames]{xcolor}

\definecolor{cvprblue}{rgb}{0.21,0.49,0.74}
\usepackage[pagebackref,breaklinks,colorlinks,citecolor=cvprblue]{hyperref}

\def\paperID{*****} 
\def\confName{CVPR}
\def\confYear{2024}

\title{Highest Score Award to the CVPR'2024 LOVEU Track1 Challenge: \\ Zero-Shot Long-Form Video Understanding through Screenplay}

\author{Yongliang Wu$^{1,3}$, Bozheng Li$^{2,3}$, Jiawang Cao$^3$, Wenbo Zhu$^3$, Yi Lu$^{3,4}$ \\ Weiheng Chi$^{3,5}$, Chuyun Xie$^{3}$, Haolin Zheng$^3$, Ziyue Su$^3$, Jay Wu$^3$, Xu Yang$^{1}$\\
$^1$Southeast University   $^2$Zhejiang University   $^3$Opus AI Research \\
$^4$The University of Manchester $^5$National University of Singapore \\
{\tt\small $\{$yongliangwu, palm\_yangxu$\}$@seu.edu.cn; }\\
{\tt\small $\{$gavin.cao, vito.zhu, sharon.xie, harley.zheng, lirian.su, jay.wu$\}$@opus.pro; }\\
{\tt\small bozhengli@zju.edu.cn; yi.lu-14@student.manchester.ac.uk; weiheng\_chi@u.nus.edu}
}

\begin{document}
\maketitle

\begin{abstract}
The Long-form Video Question-Answering task requires the comprehension and analysis of extended video content to respond accurately to questions by utilizing both temporal and contextual information. In this paper, we present MM-Screenplayer, an advanced video understanding system with multi-modal perception capabilities that can convert any video into textual screenplay representations. Unlike previous storytelling methods, we organize video content into scenes as the basic unit, rather than just visually continuous shots. Additionally, we developed a ``Look Back'' strategy to reassess and validate uncertain information, particularly targeting breakpoint mode. MM-Screenplayer achieved highest score in the CVPR'2024 LOng-form VidEo Understanding (LOVEU) Track 1 Challenge, with a global accuracy of 87.5\% and a breakpoint accuracy of 68.8\%.
\end{abstract}

\section{Introduction}
\begin{figure*}[t]
    \centering
    \includegraphics[width=\textwidth]{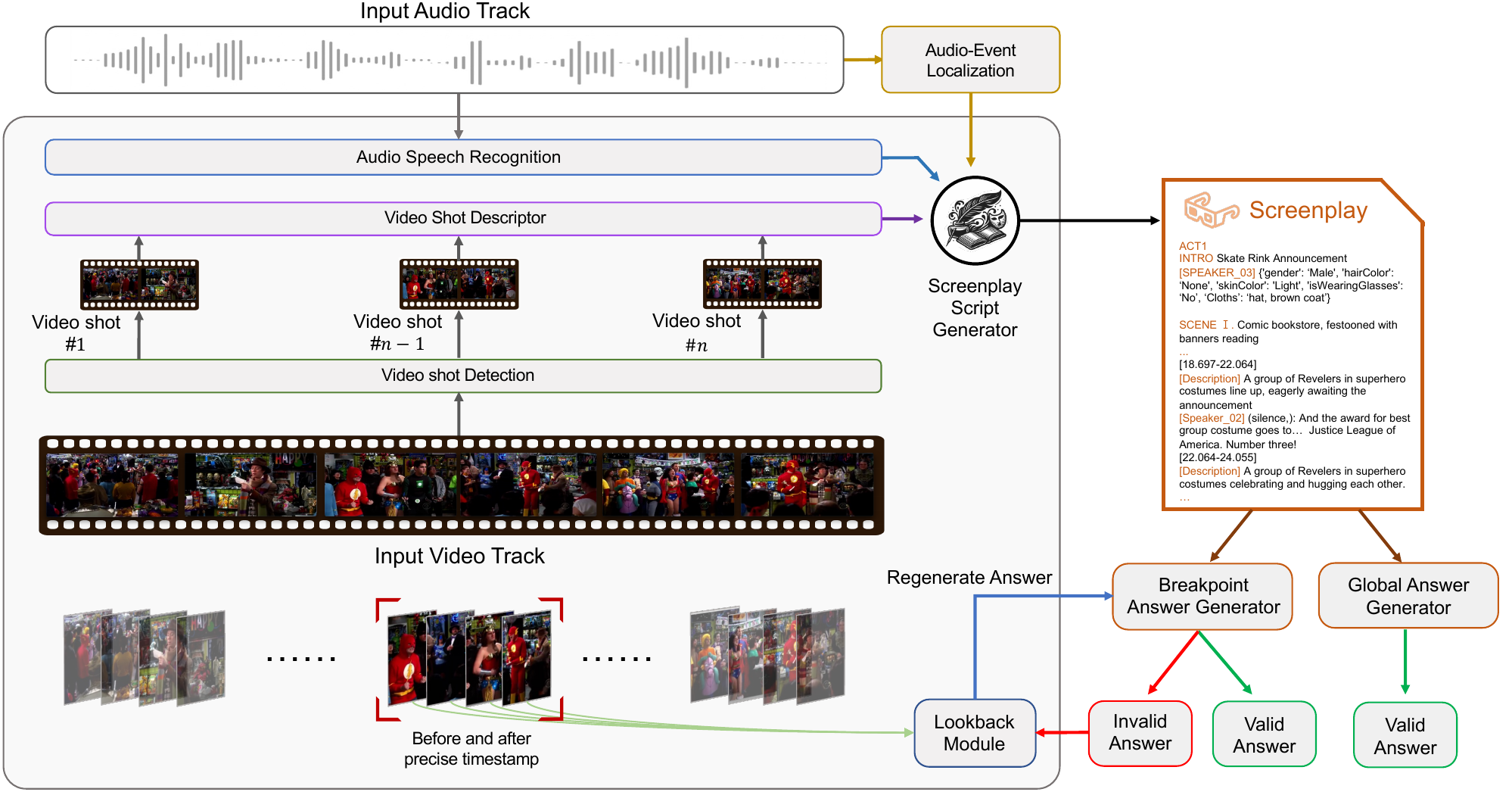}
    \caption{The overall architecture of MM-Screenplayer.}
    \vspace{-10pt}
    \label{fig:framework}
\end{figure*}

With the rapid development of video models, significant progress has been made in the domain of video understanding. However, the length of videos that these models can effectively handle remains limited. The long-form Video Question-Answering (LVQA) task has been introduced to explore the potential of models in understanding long-duration videos, specifically those videos longer than five minutes. The LVQA task demands a comprehensive understanding of the entire video from the global perspective, as well as the precise temporal capturing ability to answer questions about specific moments. This presents a substantially more challenging task in video understanding. Recently, the first long video understanding benchmark, MovieChat~\cite{song2023moviechat} has been proposed. It includes 1,000 high-quality video clips sourced from various movies and TV series, accompanied by 14,000 manual annotations. MovieChat enables quantitative evaluation of long-form video understanding capabilities in question-answering.

Most previous video models focus on end-to-end training to build a video question-answering system. Works like MovieChat~\cite{song2023moviechat, maaz2023video} rely on the question input to construct video representations and answers. Due to the lack of high-quality large-scale annotated data, these models have very limited capabilities in handling the LVQA task. Another line of work adopts a series of foundational models to convert video content into textual representations, which we refer to as storytelling methods~\cite{lin2023mm, zhang2023simple}. They leverage the powerful language understanding capabilities of LLMs to comprehend video content based on the generated scripts. However, these methods either perform captioning on each individual frame or use scene detection to segment visually continuous ``shots'' as the basic unit. These approaches overlook the temporal relationships between different segments, thus limiting their understanding of the video content. For example, multiple quick-cut shots in a movie often represent a single coherent event.

To address the aforementioned issues, we introduce MM-Screenplayer, an agent system endowed with multimodal perception capabilities for tackling LVQA tasks. It comprises three essential components: the Multi-modal Perception module, which receives inputs from both visual and audio tracks of video; the Scene-Level Script Generation Module, which prompts LLMs to reassemble and comprehend shots temporally, generating high-level semantic scenes as the basic unit; and the Look back for determination Module, which extracts, analyzes, and summarizes information from video frames before and after the specified timestamp for breakpoint mode video question answering.

MM-Screenplayer achieved first place in the CVPR 2024 LOVEU Track 1 Challenge with a global accuracy of 87.5\% and a breakpoint accuracy of 68.8\% in the MoiveChat-1k long-form video understanding dataset.

\section{Method}
Given a video $\mathcal{V}$, MM-Screenplayer generates a comprehensive screenplay $\mathcal{M}$ to thoroughly represent the content of the video. Unlike previous storytelling methods, we organize textual descriptions into higher-level semantic scenes rather than individual shots, thereby promoting a deeper comprehension of the video's narrative. Additionally, to address certain issues of the breakpoint mode, we introduce a Look-Back mechanism: if the model is uncertain to make judgments solely based on the given screenplay, it then extracts frames to utilize extra visual information to improve the understanding of the given video, therefore producing more reliable answers. The overall architecture of our model is illustrated in Figure~\ref{fig:framework}.

\subsection{Multi-Modal Perception}\label{sec:perception}
Our model comprehensively analyzes both visual and audio tracks of a video to extract rich multi-modal information. For the visual track, we start with scene detection to divide the video into distinct shots. For each shot, frames are sampled at regular intervals. A Vision-Language Model such as GPT-4V is utilized to generate image captions for each frame~\cite{yang2024exploring,wu2024glance}. For the audio track, an Automatic Speech Recognition (ASR) model is applied to transcribe the dialogue from the audio. Additionally, we utilized an Audio Event Localization model to detect and index significant audio events throughout the video. By combining these processes, we captured a structured set of data encompassing visual shots, frame descriptions, dialogue transcripts, as well as audio events. This multi-modal extraction provides a robust foundation for advanced video analysis and understanding.

\subsection{Scene-Level Scripts Generation}\label{sec:scene}
The concept of \textbf{scene} is fundamental in the design of screenplays, as it provides a higher-level semantic decision of the video. In contrast, simple storytelling methods merely rely on basic scene detection and treat individual shots as primary units~\cite{lin2023mm}. Such an approach hinders the comprehensive understanding of the narrative. \\
For example, in the film \textit{Titanic (1997)}, the sequence leading up to the ship's collision with the iceberg consists of numerous quickly shifting shots. If these shots are viewed in isolation rather than treated as part of a cohesive scene, the true narrative is lost. To tackle this issue, we propose a Scene-Level Scripts Generation module that utilizes LLMs to merge shots into coherent scenes. 

\begin{table}[t]
\centering
\caption{\small Performance comparison on the MoiveChat-1K~\cite{song2023moviechat} dataset against state-of-the-art methods. The best performance is highlighted in bold, while the second-best is underlined.}
\resizebox{\columnwidth}{!}{
\begin{tabular}{l|c@{\hskip 0.5cm}c|c@{\hskip 0.5cm}c}
\toprule
\multirow{2}{*}{Method} & \multicolumn{2}{c}{Global} & \multicolumn{2}{c}{Breakpoint}  \\
 & Accuracy & \phantom{ab}Score\phantom{ab} & Accuracy & \phantom{ab}Score\phantom{ab} \\
\midrule
\multicolumn{5}{c}{\textit{Evaluation via GPT-3.5}} \\
\midrule
Video Chat~\cite{li2023videochat} & 57.8 & 3.00 & 46.1 & 2.29 \\
Video LLaMA~\cite{zhang2023video} & 51.7 & 2.67 & 39.1 & 2.04 \\
Video-ChatGPT~\cite{maaz2023video} & 47.6 & 2.55 & 48.0 & 2.45 \\
MovieChat~\cite{song2023moviechat} & 62.3 & 3.23 & 48.3 & 2.57 \\
\midrule
\multicolumn{5}{c}{\textit{Evaluation via Gemini Pro}} \\
\midrule
MM-VID~\cite{lin2023mm}  & \underline{58.6} & 2.86 & 10.4 & 0.56  \\
LLoVi~\cite{zhang2023simple} & 58.3 & \underline{2.87} & 17.8 & 1.03 \\
MovieChat~\cite{song2023moviechat} & 55.1 & 2.78 & \underline{38.5} & \underline{1.87} \\
\midrule
\textit{Ours} & \textbf{87.5} & \textbf{4.18} & \textbf{68.8} & \textbf{3.52}  \\
\bottomrule
\end{tabular}
}
\label{tab:moviechat}
\end{table}

Initially, we arranged ASR transcripts in chronological order. For dialogues with pauses longer than 2 seconds, intervening visual content becomes significant. To address this observation, we insert a ``separator'' marker between sentences with gaps of more than 2 seconds, guiding the LLM to perform an initial rough split of the transcript. In the next phase, to integrate multi-modal information and grasp the ``subtext'', we insert captions of visual and audio events between these coarse split scenes. This creates a multi-modal text representation, allowing the LLM to reprocess and refine the separation of the content. By analyzing visual descriptions, dialogues and audio events, the module is capable of identifying logical boundaries.

\subsection{Look Back for Determination}
Our pipeline employs LLMs to understand long-form video content and answer questions across different modes using tailored prompts and generated screenplays which are reusable, enhancing efficiency in LVQA by eliminating the need for re-encoding video input for each question. 

A look-back mechanism is specifically designed for handling problems in breakpoint mode. We observed that in rare cases, such as the given problem requiring time localization with a higher precision, the screenplay alone may fail to extract sufficient information for answer generation. 

Therefore, when the response produced by the Answer Generator was detected empty or invalid (such as: containing negative keywords such as ``cannot'', ``don't know'', etc.), the model will fall back to reproduce a new answer with the incorporation of visual information. 

The visual information is obtained through the video track through frame extraction. As in breakpoint mode, the exact objective timestamp is given by the question, the pipeline will extract the frames slightly before and after that timestamp consecutively, as the retrieved information is used for producing the new answer.

By combining the extracted visual time series information with the produced screenplay, the model was able to not only gather sufficient information near the provided video breakpoint but also comprehend the video plot globally, therefore being more probable to produce valid and correct answers to breakpoint problems. 

\section{Experiments}
\subsection{Experimental Settings}
The LVQA Challenge offers 170 videos as the test set to evaluate the model's performance. The evaluation process is processed by the competition organizer using unreleased answers.
In our proposed screenplay generation pipeline, we employed GPT-4-turbo~\cite{achiam2023gpt} as the LLM driving all text script processing tasks. For visual description generation, we cherry-picked GPT-4o as the corresponding VLM. Additionally, we integrated whisperX~\cite{bain2023whisperx} as the ASR model. Gemini-1.5 pro~\cite{reid2024gemini} was chosen as the audio analyzer for audio event localization. The versions and parameters of the LLM and VLM are fixed to ensure reproducibility. All experiments are conducted on a single T4 GPU without any extra training process.
\begin{figure*}[t]
    \centering
    \includegraphics[width=\textwidth]{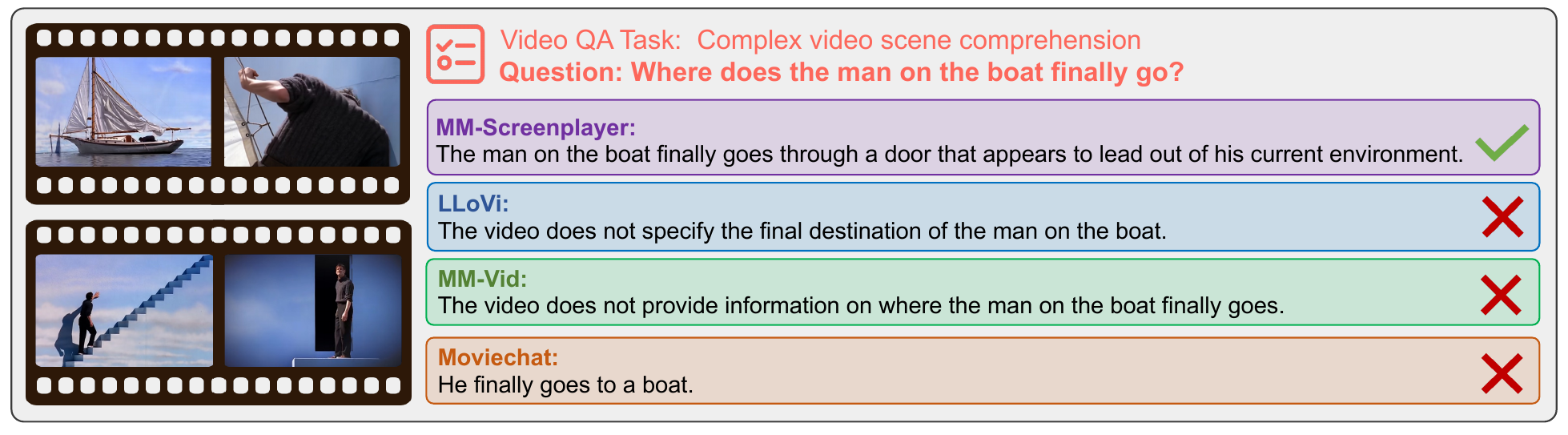}
    \caption{Comparison of answers produced by MM-Screenplayer and other state-of-the-art methods for a question from MovieChat1K-testset. Our method produced significantly better answers while all other methods' answers were incorrect.}
    \vspace{-10pt}
    \label{fig:sample}
\end{figure*}

\subsection{Main Results}
As shown in Table \ref{tab:moviechat}, MM-Screenplayer achieved top performance on the MovieChat-1K test set, with a global accuracy of 87.5\% and a global score of 4.18. In breakpoint mode, our method attained an accuracy of 68.8\% and a score of 3.52, ranking highest among all participants. 

The outstanding performance in both global and breakpoint modes demonstrates that our proposed screenplay format plays a pivotal role in representing long video content and providing rich contextual information. This enables LLM to understand long-form video content and accurately answer questions. Compared to the locally reproduced MovieChat~\cite{song2023moviechat} baseline, our pipeline generates answers that exhibit a comprehensive understanding of long-form video content and accurate temporal capture ability.

\begin{table}[t]
\centering
\caption{Performance Metrics with Different Components. Acronyms: SSGM - Scene-level Scripts Generation Module, LBDM - Look Back for Determination Module, G-Acc - Global Accuracy, G-Score - Global Score, B-Acc - Breakpoint Accuracy, B-Score - Breakpoint Score. }
\small
\begin{tabular}{c|c|c|c|c|c}
\toprule
SSGM & LBDM & G-Acc & G-Score & B-Acc & B-Score \\
\midrule
\text{\sffamily $\times$} & \text{\sffamily $\times$} & 66.7 & 3.60 & 48.5 & 2.51 \\
\text{\sffamily \checkmark} & \text{\sffamily $\times$} & 85.6 & 4.18 & 54.8 & 2.77 \\
\text{\sffamily \checkmark} & \text{\sffamily \checkmark} & 87.5 & 4.18 & 68.8 & 3.52 \\
\bottomrule 
\end{tabular}
\end{table}
\label{tab:ablation}

\subsection{Ablation Study}
We conducted extensive ablation studies on our proposed Scene-Level Scripts Generation Module and Look-Back for Determination Module as shown in Table~\ref{tab:ablation}. The results show that organizing shots into higher-level scenes as the basic unit significantly improves global accuracy. Furthermore, the introduction of the Look-Back strategy greatly enhances performance in breakpoint mode. These findings demonstrate the effectiveness of both modules. However, the results also indicate that solely relying on screenplay content is unreliable for addressing some detailed issues in breakpoint mode. This is because our visual descriptions might not have adequately captured the necessary information.

\vspace{-10pt}
\subsection{Qualititave Results}
Figure~\ref{fig:sample} presents one of our answers on MovieChat-1K test set. MM-Screenplayer produced significantly better answers which precisely captured the environmental transition of this clip of ``Truman's World'', while other methods either produced incorrect destinations or could not answer the question at all. Screenplay enables our method to accurately capture the high-level global semantics of this movie clip while the look-back mechanism could further guarantee the understanding of specific moments along with the context provided by the screenplay.

\vspace{-10pt}
\section{Conclusion}
In this paper, we introduced MM-Screenplayer, a comprehensive video understanding system with multi-modal perception capabilities that can convert videos of arbitrary length into higher-level textual representations. Our innovative approach organizes video content into scenes as the basic unit and employs a ``Look Back'' strategy to reassess and validate uncertain information. MM-Screenplayer achieved the highest score in the CVPR'2024 LOng-form VidEo Understanding (LOVEU) Track 1 Challenge, demonstrating exceptional proficiency in long-form video understanding. This accomplishment underscores the effectiveness of our method in comprehending and analyzing extended video content for accurate question-answering.

\vspace{-10pt}
{
    \small
    \bibliographystyle{ieeenat_fullname}
    \bibliography{main}

\begin{thebibliography}{11}
\providecommand{\natexlab}[1]{#1}
\providecommand{\url}[1]{\texttt{#1}}
\expandafter\ifx\csname urlstyle\endcsname\relax
  \providecommand{\doi}[1]{doi: #1}\else
  \providecommand{\doi}{doi: \begingroup \urlstyle{rm}\Url}\fi

\bibitem[Achiam et~al.(2023)Achiam, Adler, Agarwal, Ahmad, Akkaya, Aleman, Almeida, Altenschmidt, Altman, Anadkat, et~al.]{achiam2023gpt}
Josh Achiam, Steven Adler, Sandhini Agarwal, Lama Ahmad, Ilge Akkaya, Florencia~Leoni Aleman, Diogo Almeida, Janko Altenschmidt, Sam Altman, Shyamal Anadkat, et~al.
\newblock Gpt-4 technical report.
\newblock \emph{arXiv preprint arXiv:2303.08774}, 2023.

\bibitem[Bain et~al.(2023)Bain, Huh, Han, and Zisserman]{bain2023whisperx}
Max Bain, Jaesung Huh, Tengda Han, and Andrew Zisserman.
\newblock Whisperx: Time-accurate speech transcription of long-form audio.
\newblock \emph{arXiv preprint arXiv:2303.00747}, 2023.

\bibitem[Li et~al.(2023)Li, He, Wang, Li, Wang, Luo, Wang, Wang, and Qiao]{li2023videochat}
KunChang Li, Yinan He, Yi Wang, Yizhuo Li, Wenhai Wang, Ping Luo, Yali Wang, Limin Wang, and Yu Qiao.
\newblock Videochat: Chat-centric video understanding.
\newblock \emph{arXiv preprint arXiv:2305.06355}, 2023.

\bibitem[Lin et~al.(2023)Lin, Ahmed, Li, Lin, Azarnasab, Yang, Wang, Liang, Liu, Lu, et~al.]{lin2023mm}
Kevin Lin, Faisal Ahmed, Linjie Li, Chung-Ching Lin, Ehsan Azarnasab, Zhengyuan Yang, Jianfeng Wang, Lin Liang, Zicheng Liu, Yumao Lu, et~al.
\newblock Mm-vid: Advancing video understanding with gpt-4v (ision).
\newblock \emph{arXiv preprint arXiv:2310.19773}, 2023.

\bibitem[Maaz et~al.(2023)Maaz, Rasheed, Khan, and Khan]{maaz2023video}
Muhammad Maaz, Hanoona Rasheed, Salman Khan, and Fahad~Shahbaz Khan.
\newblock Video-chatgpt: Towards detailed video understanding via large vision and language models.
\newblock \emph{arXiv preprint arXiv:2306.05424}, 2023.

\bibitem[Reid et~al.(2024)Reid, Savinov, Teplyashin, Lepikhin, Lillicrap, Alayrac, Soricut, Lazaridou, Firat, Schrittwieser, et~al.]{reid2024gemini}
Machel Reid, Nikolay Savinov, Denis Teplyashin, Dmitry Lepikhin, Timothy Lillicrap, Jean-baptiste Alayrac, Radu Soricut, Angeliki Lazaridou, Orhan Firat, Julian Schrittwieser, et~al.
\newblock Gemini 1.5: Unlocking multimodal understanding across millions of tokens of context.
\newblock \emph{arXiv preprint arXiv:2403.05530}, 2024.

\bibitem[Song et~al.(2023)Song, Chai, Wang, Zhang, Zhou, Wu, Guo, Ye, Lu, Hwang, et~al.]{song2023moviechat}
Enxin Song, Wenhao Chai, Guanhong Wang, Yucheng Zhang, Haoyang Zhou, Feiyang Wu, Xun Guo, Tian Ye, Yan Lu, Jenq-Neng Hwang, et~al.
\newblock Moviechat: From dense token to sparse memory for long video understanding.
\newblock \emph{arXiv preprint arXiv:2307.16449}, 2023.

\bibitem[Wu and Yang(2024)]{wu2024glance}
Yongliang Wu and Xu Yang.
\newblock A glance at in-context learning.
\newblock \emph{Frontiers of Computer Science}, 18\penalty0 (5):\penalty0 185347, 2024.

\bibitem[Yang et~al.(2024)Yang, Wu, Yang, Chen, and Geng]{yang2024exploring}
Xu Yang, Yongliang Wu, Mingzhuo Yang, Haokun Chen, and Xin Geng.
\newblock Exploring diverse in-context configurations for image captioning.
\newblock \emph{Advances in Neural Information Processing Systems}, 36, 2024.

\bibitem[Zhang et~al.(2023{\natexlab{a}})Zhang, Lu, Islam, Wang, Yu, Bansal, and Bertasius]{zhang2023simple}
Ce Zhang, Taixi Lu, Md~Mohaiminul Islam, Ziyang Wang, Shoubin Yu, Mohit Bansal, and Gedas Bertasius.
\newblock A simple llm framework for long-range video question-answering.
\newblock \emph{arXiv preprint arXiv:2312.17235}, 2023{\natexlab{a}}.

\bibitem[Zhang et~al.(2023{\natexlab{b}})Zhang, Li, and Bing]{zhang2023video}
Hang Zhang, Xin Li, and Lidong Bing.
\newblock Video-llama: An instruction-tuned audio-visual language model for video understanding.
\newblock In \emph{Proceedings of the 2023 Conference on Empirical Methods in Natural Language Processing: System Demonstrations}, pages 543--553, 2023{\natexlab{b}}.

\end{thebibliography}
}

\end{document}